\documentclass[11pt]{article}

\usepackage[final]{acl}

\usepackage{times}
\usepackage{latexsym}

\usepackage[T1]{fontenc}

\usepackage[utf8]{inputenc}

\usepackage{microtype}

\usepackage{inconsolata}

\usepackage{graphicx}
\usepackage{enumitem}
\usepackage{subcaption}
\usepackage{xspace}
\usepackage{adjustbox}
\usepackage{booktabs}
\usepackage{MnSymbol}
\usepackage{multirow}
\usepackage{makecell}
\usepackage{tablefootnote}
\usepackage[table,xcdraw]{xcolor}
\usepackage{colortbl}
\usepackage{algpseudocode}
\usepackage{algorithm}
\usepackage{caption}
\usepackage{amsfonts}
\usepackage{graphicx}
\usepackage{pifont}
\usepackage{xcolor}
\usepackage{fontawesome}
\usepackage{makecell}
\usepackage{caption}
\usepackage{tikz}
\usetikzlibrary{calc}
\usepackage{booktabs}
\usepackage{pifont}

\newcommand{\cmark}{\textcolor{green}{\ding{51}}}
\newcommand{\xmark}{\textcolor{red}{\ding{55}}}   

\title{\method: Text-attributed Graph Dataset Distillation via Coupling Language Model with Graph-Aware Kernel}

\author{
\textbf{Yeongho Kim} \quad
\textbf{Yeonje Choi} \quad
\textbf{Kijung Shin} \\
Kim Jaechul Graduate School of AI, KAIST \\
\texttt{\{yeongho, yeonjechoi, kijungs\}@kaist.ac.kr}  \\
}

\begin{document}

\newcommand{\method}{\textsc{TaLK}\xspace}
\newcommand{\smallsection}[1]{{\noindent{\bf{\smash{#1}}}}}
\newcommand{\uarrow}{\small{~($\uparrow$)}}
\newcommand{\darrow}{\small{~($\downarrow$)}}
\newcommand{\cmt}[1]{\textcolor{red}{#1}}
\newcommand{\tbe}[1]{\textcolor{blue}{#1}}

\newcommand\red[1]{\textcolor{red}{#1}}
\newcommand\blue[1]{\textcolor{blue}{#1}}
\newcommand\green[1]{\textcolor{green}{#1}}
\newcommand{\yj}[1]{\textcolor{orange}{#1}}
\newcommand{\yh}[1]{\textcolor{brown}{#1}}

\newcommand\sblue[1]{\small\textcolor{blue}{#1}}
\newcommand\sred[1]{\small\textcolor{red}{#1}}

\maketitle

\begin{abstract}
Text-attributed graphs (TAGs) are widely used in many real-world domains, and learning on TAGs requires jointly modeling text semantics and graph structure. A standard approach for modeling TAGs is to combine a language model (LM) and a graph neural network (GNN), but joint training is computationally expensive and difficult to scale. Dataset distillation is a promising way to reduce training costs, 
but existing methods are not well suited to TAGs because they are typically designed for a single modality or still require repeatedly training expensive LM--GNN models on the full dataset during distillation.
To address this, we propose \method, an effective dataset distillation method for TAGs that couples an LM with a graph-aware neural tangent kernel.
This design enables efficient dataset distillation, avoiding repeated joint training on the full dataset while reflecting both textual and structural information for effective TAG learning.
Experiments on multiple TAG benchmarks show that \method consistently outperforms existing baselines and achieves up to 97\% of full-dataset performance with only 1\% synthetic data. Our code is available at \url{https://github.com/thisis05/TaLK}.
\end{abstract}

\section{Introduction}
\label{sec:intro}
Text-attributed graphs (TAGs) are widely used in many real-world domains, including academic networks~\cite{hu2020open}, social media~\cite{ma2018rumor}, and e-commerce systems~\cite{he2016ups}. In TAGs, each node is associated with rich textual content (e.g., paper abstracts or product descriptions), and edges represent meaningful relationships between nodes (e.g., citations or co-purchases).

A widely adopted approach for modeling TAGs is to combine pretrained language models (LMs) with graph neural networks (GNNs)~\cite{bi2021leveraging}, thereby integrating text semantics with graph topology~\cite{yan2023comprehensive}. Specifically, an LM encodes node texts into representations, which are then fed into a GNN for neighborhood aggregation, and the two modules are optimized jointly in an end-to-end manner.

However, the joint training of such LM–GNN integrated models does not scale well to many real-world TAGs.
A key bottleneck is that updating the LM parameters requires retaining the LM computations for neighboring nodes along with the intermediate states for trainable GNN neighborhood aggregation.
This overhead is particularly severe in many real-world TAGs, where nodes often have many neighbors.
Accordingly, this has inspired decoupled training strategies, where the LM is first optimized on node texts, and the resulting text representations are subsequently used for GNN training~\cite{duan2023simteg, wang2024bridging}.
However, they may not fully match the benefits of joint training and can still be costly at scale, especially when repeated training is required for model selection or hyperparameter tuning.

Dataset distillation offers a promising approach to alleviating such challenges by reducing the training costs of deep learning models~\cite{wang2018dataset,kwon2026effective}. 
It learns a compact set of synthetic samples that capture essential training information, enabling models trained on them to achieve performance close to that of models trained on the full dataset. Training on such synthetic datasets enables efficient hyperparameter and architecture search~\cite{yu2023dataset}.

However, dataset distillation for TAGs remains limited. While dataset distillation has been explored in NLP and graph learning~\cite{sachdeva2023data},
existing methods  designed for a single modality are not suitable for TAG learning, 
which requires leveraging both text semantics and graph structure. For TAGs, dataset distillation can benefit from modeling diverse data types jointly rather than treating them separately. Furthermore, directly applying existing dataset distillation methods to TAGs, such as gradient matching, trajectory matching, and distribution matching, remains challenging. While their objectives differ, they all repeat joint training on the full dataset to extract matching signals (e.g., gradients, training trajectories, or representation statistics) during distillation. As a result, for the same reasons discussed above, they inherit the same scalability bottleneck.

To address these limitations, we propose \textbf{\method} (\textbf{T}ext-\textbf{a}ttributed graph dataset distillation via coupling \textbf{L}anguage model with graph-aware \textbf{K}ernel), a novel dataset distillation method for TAGs. \method distills the synthetic dataset under the joint LM–GNN training pipeline, so that both text semantics and graph structure are captured during distillation. For efficient dataset distillation under the joint training pipeline, we integrate an LM encoder with the graph-aware neural tangent kernel~\cite{du2019graph,wang2024fast}, which incorporates structural information in kernel space without using a trainable GNN to aggregate neighborhood information.

Specifically, we parameterize the synthetic dataset as a small set of learnable token embeddings in a continuous input space and optimize it via a bi-level procedure. In the inner-loop, an LM and a GNN are jointly trained on the small synthetic dataset. In the outer-loop, we optimize the synthetic dataset via kernel ridge regression with a graph-aware kernel built on LM representations.
In addition, we introduce \textit{batch-wise gradient injection}, which enables scalable mini-batch training by making mini-batch optimization compatible with the kernel-based objective. This design eliminates the need for repeated joint training of LM–GNN integrated models on the full dataset during distillation, thereby improving scalability.

Through extensive experiments on multiple TAG benchmarks, we demonstrate the strengths of \method, which are summarized as follows:
\begin{itemize}[leftmargin=*]
    \item \textbf{Efficient:} 
    To the best of our knowledge, \method is the first method that efficiently enables dataset distillation directly under joint LM–GNN training, without decoupling, in practical settings. 
    
    \item \textbf{Effective:} 
    \method consistently outperforms existing baselines across multiple TAG benchmarks. Notably, it achieves up to 97\% of full-dataset performance with only 1\% synthetic data.

    \item \textbf{Practical:} We further demonstrate that the synthetic datasets from \method enable efficient hyperparameter search and transfer well across different architectures. 
    
\end{itemize}

\section{Related Works}
\label{sec:related}
\subsection{Text-attributed Graph (TAG) Learning}

As discussed in Section~\ref{sec:intro}, despite the effectiveness of joint LM--GNN training for TAG learning, many works adopt a decoupled training strategy due to its computational and memory costs.
A representative example is a two-stage training pipeline, where an LM is first trained to produce informative representations of node texts, and a GNN is subsequently trained separately using these representations as fixed input features~\cite{duan2023simteg, he2023harnessing}.
Some works follow this two-stage pipeline but incorporate graph-derived context, such as neighborhood texts, into the LM training before the GNN training stage~\cite{chien2022node, wang2024bridging}.

Several methods extend the simple two-stage training by enabling interaction between separately trained LM and GNN components through iterative optimization or by aligning the representations~\cite{zhao2022learning, li2023grenade}.
Another line of research encourages the LM to capture structural information without relying on separate GNN training, for example, through pre-training with relation prediction or structure-aware text augmentation~\cite{yasunaga2022linkbert, zhou2025taming}.

In this work, we focus on a standard LM--GNN architecture for generality and aim to make its joint training feasible through dataset distillation, without decoupling the LM and the GNN.

\subsection{Graph and Text Dataset Distillation}

Dataset distillation~\cite{wang2018dataset} aims to condense a large training dataset into a small synthetic dataset such that models trained on the synthetic dataset achieve performance comparable to those trained on the full training dataset. 
It has been extensively studied for graphs~\cite{gonggc4nc}, and representative approaches include gradient matching~\cite{jin2021graph,yang2023does}, kernel-based prediction matching~\cite{wang2024fast}, trajectory matching~\cite{zheng2023structure,zhang2024navigating}, and distribution matching~\cite{liu2022graph}. 

In parallel, text dataset distillation methods have been explored, including embedding-level performance 
matching~\cite{li2021data,maekawa2023dataset}, generator-based approaches~\cite{tao2024textual,maekawa2024dilm}, and LLM-driven approaches~\cite{shen2025condenselm}.

Despite this progress, applying dataset distillation methods for graphs and text separately is suboptimal for TAGs, which require jointly capturing text semantics and graph structure. We therefore develop a more effective approach for TAGs.

\section{Preliminaries}
\label{sec:prelim}
\subsection{Basic Concepts}
\label{sec:prelim:concepts}

\paragraph{Text-attributed graph.}
A text-attributed graph (TAG) with $N$ nodes is defined as $\mathcal{T}=(\mathbf{X},\mathbf{A})$, where $\mathbf{X}=\{x_i\}_{i=1}^{N}$ denotes textual attributes
associated with each node,
and $\mathbf{A}\in\{0,1\}^{N\times N}$ is the adjacency matrix with $A_{ij}=1$ if nodes $i$ and $j$ are adjacent and $A_{ij}=0$ otherwise, representing structural relations between nodes.
A citation network where papers are nodes and their abstracts serve as textual attributes is a typical example of a TAG.

\paragraph{Node classification.}
Let $L$ denote the set of labeled nodes. For each node $i\in L$, $y_i\in C$ denotes its label, where $C$ is the set of classes.
Semi-supervised node classification aims to predict the labels of the remaining unlabeled nodes, and it is a representative downstream task for TAG learning.

\subsection{Text-attributed Graph  Learning\label{sec:prelim:pipeline}}

TAG learning learns node representations by leveraging both textual attributes 
$\mathbf{X}$ and graph structure $\mathbf{A}$.
A widely adopted architecture for this purpose is an integrated LM–GNN model.
The LM encoder $f_{\phi}$ first maps node texts $\mathbf{X}$ to semantic representations of dimension $F$:
\begin{equation}
    \mathbf{H}=f_{\phi}(\mathbf{X})\in\mathbb{R}^{N\times F},
\end{equation}
where each row of $\mathbf{H}$ is a pooled LM output (e.g., the \texttt{[CLS]} token embedding). Then, the GNN $g_{\psi}$ refines these representations through neighborhood aggregation over $\mathbf{A}$, to capture relational structure,
and produces node-level predictions
\begin{equation}
\mathbf{Z}=g_{\psi}(\mathbf{A},f_{\phi}(\mathbf{X}))\in \mathbb{R}^{N\times|C|}.
\end{equation}
Such models are typically trained either via \emph{joint training} or \emph{decoupled training}, as described below.

\paragraph{Joint training.}
In the joint training pipeline, the LM and the GNN are optimized simultaneously in an end-to-end manner:
\begin{equation}
    \min_{\phi,\psi}\frac{1}{|L|}\sum_{i\in L}\mathcal{L}_{\mathrm{CE}}\!\left(g_{\psi}(\mathbf{A},f_{\phi}(\mathbf{X}))_i, y_i\right),
    \label{eq:joint}
\end{equation}
where $\mathcal{L}_{\mathrm{CE}}(\cdot,\cdot)$ denotes the cross-entropy loss for node classification.

\paragraph{Decoupled training.} \label{sec:prelim:decoupled}
As discussed in Section~\ref{sec:intro}, due to the high computational and memory costs of joint training, a decoupled two-stage pipeline has been widely considered as a practical alternative.
In the first stage, the LM is optimized with a task head $W$ for node classification:
\begin{equation}
    \min_{\phi,W}\frac{1}{|L|}\sum_{i\in L}\mathcal{L}_{\mathrm{CE}}\!\left((f_{\phi}(\mathbf{X})W)_i, y_i\right).
    \label{eq:stage1_decoupled}
\end{equation}
In the second stage, the trained LM generates node representations $\mathbf{H}$, which are treated as fixed initial node features for training the subsequent GNN:
\begin{equation}
    \min_{\psi}\frac{1}{|L|}\sum_{i\in L}\mathcal{L}_{\mathrm{CE}}\!\left(g_{\psi}(\mathbf{A},\mathbf{H})_i, y_i\right).
    \label{eq:stage2_decoupled}
\end{equation}

\section{Problem: Dataset Distillation for TAGs}
\label{sec:distillation}
\begin{figure*}
    \centering
        \includegraphics[width=\linewidth]{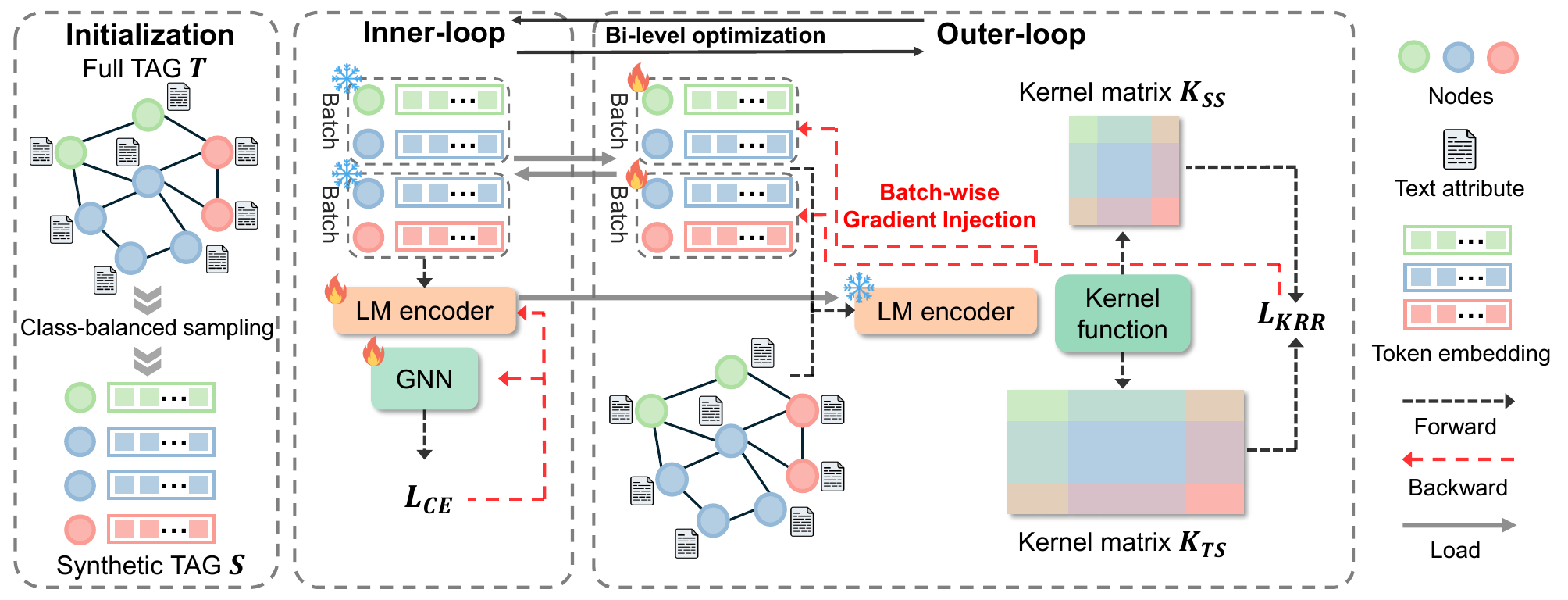}
        \caption{
        Overview of the distillation stage of \method from initialization to bi-level optimization. The inner-loop performs joint LM--GNN training on the synthetic dataset, while the outer-loop updates the synthetic dataset using a kernel-based objective through batch-wise gradient injection.
        \label{fig:model}}
\end{figure*}

In this section, we first briefly review the standard dataset distillation framework that serves as the basis of our method~\cite{wang2018dataset}. We then describe the key challenges in designing a dataset distillation method for TAGs.

\subsection{Standard Dataset Distillation\label{sec:distillation:dd}}

Given a full dataset $\mathcal{D}$ with $N$ samples, dataset distillation aims to construct a synthetic dataset $\mathcal{S}$ with $M$ samples, where $M \ll N$. 
Dataset distillation is formulated as a bi-level optimization problem. First, the inner-loop aims to optimize model parameters (e.g., $\phi$ and $\psi$ in Eq.~\eqref{eq:joint}) using $\mathcal{S}$. Starting from the initial model parameters $\theta_0$, the inner-loop typically computes, for $t = 0,1,\ldots,T-1$, 
\begin{equation}
    \theta_{t+1}
    = 
    \theta_{t}-\eta\,\nabla_{\theta_{t}}\mathcal{L}\big(\theta_{t};\mathcal{S}\big),
    \label{eq:inner_dd}
\end{equation}
where $\eta$ is a learning rate, and $\mathcal{L}(\theta_{t};\mathcal{S})$ denotes the task loss of the model with parameters $\theta_{t}$ on $\mathcal{S}$. The outer-loop aims to optimize $\mathcal{S}$ to minimize the loss on $\mathcal{D}$ after the inner-loop optimization:
\begin{equation}
    \min_{\mathcal{S}}\ \mathcal{L}\Big(\theta_{T}(\mathcal{S});\mathcal{D}\Big), 
    \label{eq:outer_dd}
\end{equation}
where $\theta_{T}(\mathcal{S})$ denotes the model parameters $\theta_{T}$ trained on $\mathcal{S}$. By iterating this bi-level optimization, $\mathcal{S}$ is optimized such that a model trained on $\mathcal{S}$ achieves strong performance on the full dataset $\mathcal{D}$.

\subsection{Challenges in TAG Dataset Distillation}\label{sec:distillation:problems}
TAG learning can benefit from dataset distillation, as it incurs substantial computational and memory costs. Effective dataset distillation for TAGs should remain computationally feasible while jointly considering text semantics and graph structure, rather than treating them separately. However, applying existing approaches does not readily satisfy both requirements. We discuss these challenges below.

\paragraph{Existing methods can be costly for TAGs.}\label{sec:distillation:problems:p1}
First, existing dataset distillation methods incur high computational and memory costs when applied to TAGs.
In the outer-loop optimization of Eq.~\eqref{eq:outer_dd}, the synthetic dataset is optimized by backpropagating the loss evaluated on the full dataset.
As discussed in Section~\ref{sec:intro}, this is expensive for TAGs in terms of both time and memory, especially under joint training, as it requires repeatedly training an LM–GNN integrated model. 
This limitation applies to various dataset distillation methods, including gradient matching and trajectory matching.

\paragraph{Decoupled distillation is feasible but suboptimal.}\label{sec:distillation:problems:p2}
An alternative is to distill text and graph separately under the decoupled training pipeline in Section~\ref{sec:prelim:decoupled}. While computationally more feasible, this approach is limited as it treats text semantics and graph structure independently rather than jointly. Since effective TAG learning requires integrating both modalities, such independent distillation may yield suboptimal synthetic data.

\section{Proposed Method: \method}
\label{sec:method}

\begin{algorithm}[t]
    \caption{Distillation stage of \method.}
    \label{alg:distill}
    \begin{algorithmic}[1]
        \Statex \textbf{Input:} Full TAG dataset $\mathcal{T}$ with node labels $\mathbf{Y}_L$; pretrained LM parameters $\phi^{(0)}$; inner and outer learning rates $\eta$ and  $\tilde{\eta}$; epochs $K$; inner and outer-loop iteration steps $T_{\text{in}}$ and $T_{\text{out}}$.
        \State Initialize synthetic dataset $S_0=(\mathbf{X'_{0}},\mathbf{I})$ and node labels $\mathbf{Y'}$
        \For{$k=0$ \textbf{to} $K-1$}
            \State Initialize LM parameters $\phi_0$ to $\phi^{(0)}$
            \State Initialize 
            GNN parameters $\psi_0$
            \State Set $\theta_0 \leftarrow (\phi_0,\psi_0)$
            \For{$t=0$ \textbf{to} $T_{\text{in}}-1$}
                \State $\theta_{t+1} \leftarrow \theta_t - \eta\,\nabla_{\theta_t}\mathcal{L}_{\mathrm{CE}}(\theta_t;\mathcal{S}_k)$ \hfill {\footnotesize (Eq.~\eqref{eq:joint_lm_gnn_general})}
            \EndFor
            \For{$T_{\text{out}}$ steps}
                \State $\mathbf{X}'_{k} \leftarrow \mathbf{X}'_{k} - \tilde{\eta}\,\nabla_{\mathbf{X}'_{k}}\mathcal{L}_{\mathrm{KRR}}$ \hfill {\footnotesize (Eq.~\eqref{eq:krr_obj}--\eqref{eq:batch_gradient})}
            \EndFor
            \State $\mathcal{S}_{k+1} \leftarrow (\mathbf{X}'_{k},\mathbf{I})$
        \EndFor
        \Statex \textbf{Output:} $\mathcal{S}_K$
    \end{algorithmic}
\end{algorithm}

In this section, we introduce \method, a novel dataset distillation method for text-attributed graphs (TAGs). 
\method retains the joint LM--GNN training pipeline to preserve distillation effectiveness, while replacing the expensive outer-loop with a memory-efficient kernel-based objective to reduce computational and memory overhead.

\subsection{Overview}

\paragraph{Distillation stage.} We distill the original text-attributed graph dataset $\mathcal{T}$ into a synthetic dataset $\mathcal{S}$ using a pretrained LM and the graph-aware neural tangent kernel within a bi-level optimization framework.
Specifically, we first initialize $\mathcal{S}$ by 
sampling from $\mathcal{T}$, and train an LM and a GNN on $\mathcal{S}$ in the inner-loop (Section~\ref{sec:method:inner}). In the outer-loop, we replace explicit GNN training with a graph-aware neural tangent kernel and optimize $\mathcal{S}$ via a kernel ridge regression objective so that the predictions induced by the synthetic dataset match those of the full dataset (Section~\ref{sec:method:outer}). Furthermore, we propose \textit{batch-wise gradient injection} to apply the graph-aware neural tangent kernel effectively (Section~\ref{sec:method:bgi}). The overall distillation process is described in Figure~\ref{fig:model} and Algorithm~\ref{alg:distill}.

\paragraph{Training and inference stage.} After dataset distillation, an LM and a GNN are jointly trained on $\mathcal{S}$. 
The resulting LM–GNN integrated model is then used for inference, typically on the full dataset $\mathcal{T}$.

\subsection{Initialization and Inner-loop}\label{sec:method:inner}

\paragraph{Initialization.} We first initialize the synthetic dataset $\mathcal{S}=(\mathbf{X}',\mathbf{A}')$ with associated node labels $\mathbf{Y}'$ by sampling $M$ nodes from the full training dataset using class-balanced sampling so that the label distribution of $\mathbf{Y}'$ matches that of the original training labels. 
We then obtain the initial synthetic features $\mathbf{X}'$ by passing the textual attributes of the sampled nodes through the word embedding layer of a pretrained LM to form token-level embeddings. 
Following the structure-free graph dataset distillation paradigm~\cite{zheng2023structure,zhang2024navigating}, we fix the synthetic structure as an identity matrix, i.e., $\mathbf{A}'=\mathbf{I}$. 
The key intuition is that, during distillation, the optimized synthetic features can implicitly capture the structural information of the original graph, thereby alleviating the need to explicitly synthesize an adjacency matrix (see Appendix~\ref{sec:app:sfdd} for details). 
Accordingly, $\mathcal{S}=(\mathbf{X}',\mathbf{A}')$ is initialized to $\mathcal{S}_{0}=(\mathbf{X}'_{0},\mathbf{I})$, where $\mathbf{X}'_{0}\in\mathbb{R}^{M\times Q\times F}$, with $M$ denoting the number of sampled nodes and $Q$ the maximum token sequence length. 
We treat $\mathbf{Y}'$ as fixed hard labels inherited from the sampled nodes, and only optimize $\mathbf{X}'$ during distillation.

\paragraph{Inner-loop.} In the inner-loop, we jointly train the pretrained LM encoder $f_{\phi}$ and a GNN $g_{\psi}$ on the synthetic dataset $\mathcal{S}=(\mathbf{X}',\mathbf{I})$ with labels $\mathbf{Y}'$:
\begin{equation}
    \min_{\phi,\psi}\ \mathcal{L}_{\mathrm{CE}}\!\left(g_{\psi}\!\left(\mathbf{I}, f_{\phi}(\mathbf{X}')\right),\mathbf{Y}'\right),
    \label{eq:joint_lm_gnn_general} 
\end{equation}
where $\mathcal{L}_{\mathrm{CE}}$ denotes the cross-entropy loss for node classification. The use of GNN $g_{\psi}$ in the inner-loop ensures that the LM parameters $\phi$ are updated by gradients flowing through the GNN, aligning with the joint training pipeline. 

\subsection{Outer-loop Optimization}\label{sec:method:outer}

\paragraph{Overview of outer-loop.} As discussed in Section~\ref{sec:distillation:problems:p1}, optimizing $\mathcal{S}$ in the outer-loop (Eq.~\eqref{eq:outer_dd}) is challenging due to its substantial computational and memory costs. To mitigate this bottleneck, we extend a prior neural tangent kernel-based  approach~\cite{nguyen2020dataset} to TAG dataset distillation. Specifically, in the outer-loop, we replace the GNN with a graph-aware neural tangent kernel, which captures structural information via fixed kernel evaluations, without requiring GNN training. 
This substitution substantially improves efficiency while preserving structural information, and enables label prediction via the closed-form solution of kernel ridge regression.
Note that this substitution is restricted to the outer-loop, and joint LM–GNN training is preserved in the inner-loop through the explicit use of a GNN.
\paragraph{Kernel function.}
To define the outer-loop objective, we first explain a node-level graph-aware neural tangent kernel $\mathbf{K}(\cdot,\cdot)$. Given text-derived node representations as inputs, $\mathbf{K}(\cdot,\cdot)$ computes pairwise kernel values between nodes while reflecting graph structure. Specifically, using the  LM encoder $f_{\phi}$ trained in the inner-loop, we obtain node representations on the full dataset $\mathcal{T}$ and the synthetic dataset $\mathcal{S}$:
\begin{equation}\label{eq:LMrep}
\mathbf{H}_{\mathcal{T}} = f_{\phi}(\mathbf{X}) \in \mathbb{R}^{N \times F}, \ \mathbf{H}_{\mathcal{S}} = f_{\phi}(\mathbf{X}') \in \mathbb{R}^{M \times F},
\end{equation}
and based on them, obtain kernel matrices
\begin{align}
\mathbf{K}_{\mathcal{S}\mathcal{S}} & = \mathbf{K}(\mathbf{H}_{\mathcal{S}}, \mathbf{H}_{\mathcal{S}})\in \mathbb{R}^{M\times M}, \label{eq:kerenl_form1}\\
\mathbf{K}_{\mathcal{T}\mathcal{S}} & = \mathbf{K}(\mathbf{H}_{\mathcal{T}}, \mathbf{H}_{\mathcal{S}})\in \mathbb{R}^{N\times M},\label{eq:kerenl_form2}
\end{align}

which denote the kernel matrices over $\mathcal{S}$ and the cross-kernel matrix between $\mathcal{T}$ and $\mathcal{S}$, respectively. Specifically, we adopt the structure-based neural tangent kernel (SNTK)~\cite{wang2024fast} for $\mathbf{K}(\cdot,\cdot)$, which incorporates local neighborhood aggregation into the neural tangent kernel to reflect graph structure, and adapt it efficiently for TAGs, 
as summarized below (see Appendix~\ref{sec:app:implement:sntk} for details).
\begin{itemize}[leftmargin=*]
    \item \textbf{Efficient neighborhood aggregation.} The original SNTK constructs a large Kronecker-product matrix for aggregation, incurring high memory overhead. We instead compute the same aggregation directly using the adjacency and kernel matrices, avoiding explicit construction.
    \item \textbf{Efficient approximation for normalization.} The original implementation maintains full self-kernel matrices to obtain normalization terms. We instead use a feature-based approximation of these normalization terms, eliminating the need for retaining full self-kernel matrices and further improving memory efficiency in practice.
\end{itemize}

\paragraph{Kernel ridge regression.}
Using these kernel matrices, we formulate the outer-loop objective via kernel ridge regression (KRR), which yields a closed-form label predictor in kernel space. Given the synthetic set $\mathcal{S}$, the KRR prediction of node labels on the full dataset $\mathcal{T}$ is obtained by
\begin{equation}
\label{eq:krr_obj}
    \hat{\mathbf{Y}} =
    \mathbf{K}_{\mathcal{T}\mathcal{S}}
    \left(\mathbf{K}_{\mathcal{S}\mathcal{S}} + \lambda \mathbf{I}\right)^{-1}
    \mathbf{Y}',
\end{equation}
where $\lambda>0$ is the ridge regularizer. The outer-loop loss is defined on the labeled nodes:
\begin{equation}
\label{eq:krr_loss_labeled}
    \mathcal{L}_{\mathrm{KRR}}
    =
    \frac{1}{2}
    \left\|
    \mathbf{Y}_{L} - \hat{\mathbf{Y}}_{L}
    \right\|_F^2.
\end{equation}
That is, we optimize $\mathcal{S}$ by minimizing the prediction error of the KRR predictor 
from $\mathcal{S}$ over the labeled nodes on $\mathcal{T}$.
This enables $\mathcal{S}$ to encode both relevant text semantics and structural information.

\subsection{Batch-wise Gradient Injection}\label{sec:method:bgi}
Despite the improved efficiency of the outer-loop, computing the kernel function on top of the LM encoder still incurs a high memory cost, making full-batch training challenging. While adopting standard mini-batch training is feasible, it does not faithfully preserve the original objective. The outer-loop objective in Eq.~\eqref{eq:krr_obj}--\eqref{eq:krr_loss_labeled} depends on the global context of the synthetic dataset through both $\mathbf{K}_{\mathcal{T}\mathcal{S}}$ and $\mathbf{K}_{\mathcal{S}\mathcal{S}}$. Consequently, naively partitioning $\mathbf{X}'$ into mini-batches and optimizing them batch by batch would change the underlying optimization problem itself. Specifically, instead of learning a single and cohesive synthetic dataset of size $M$, 
the outer-loop would essentially optimize multiple smaller synthetic datasets in isolation.

To preserve the original outer-loop objective while enabling scalable mini-batch training, we propose \textit{batch-wise gradient injection}. Let the synthetic set be partitioned into $B$ batches, denoted by $\mathcal S^{(b)} = (\mathbf X'^{(b)}, \mathbf I^{(b)})$ with associated labels $\mathbf{Y}'^{(b)}$, where $b=1,\dots,B$. 
We first encode the synthetic batches through the LM and concatenate them to form the global representation $\mathbf{H}_{\mathcal{S}}$. The outer-loop loss $\mathcal{L}_{\mathrm{KRR}}$ is then calculated using this global representation. From this objective, the gradient with respect to $\mathbf{H}_{\mathcal{S}}$ can be computed as:
\begin{equation}
    \mathbf G
    \;\triangleq\;
    \nabla_{\mathbf H_{\mathcal S}} \mathcal L_{\mathrm{KRR}}.
\end{equation}
We then partition $\mathbf{G}$ into batch-specific slices $\{\mathbf{G}^{(b)}\}_{b=1}^B$ and inject each $\mathbf{G}^{(b)}$ as the upstream gradient for its corresponding batch. Subsequent backpropagation is then performed through the LM for each batch, yielding gradients with respect to the synthetic features $\mathbf{X}'^{(b)}$ as follows:
\begin{equation}
\label{eq:batch_gradient}
    \nabla_{\mathbf {X}'^{(b)}} \mathcal L_{\mathrm{KRR}}
    \;=\;
    \left(
    \frac{\partial f_\phi(\mathbf{X}'^{(b)})}{\partial \mathbf{X}'^{(b)}}
    \right)^{\!\top}
    \mathbf G^{(b)}.
\end{equation}
Although gradients are propagated through the LM in a batch-wise manner, each $\mathbf G^{(b)}$ is derived from the outer-loop loss evaluated on the global synthetic dataset, and therefore already reflects the global interactions induced by the kernel function and KRR. In this way, we preserve the original outer-loop objective defined on the full synthetic dataset while performing LM backpropagation in mini-batches.

\section{Experimental Results}
\label{sec:exp}
\newcolumntype{C}[1]{>{\centering\arraybackslash}m{#1}}
\newcommand{\accstd}[2]{\scalebox{0.95}{#1}{\scalebox{0.7}{$\pm$#2}}}

\begin{table*}[t]
\centering
\small
\setlength{\tabcolsep}{1.2pt}
\renewcommand{\arraystretch}{1.08}

\begin{tikzpicture}
\node[inner sep=0pt, outer sep=0pt] (tbl) {%
\begin{tabular}{@{}C{1.3cm}C{0.78cm}ccccccC{1.2cm}ccccc@{}}
\toprule
\multirow{4}{*}[-0.65em]{\shortstack{Dataset}} 
& \multirow{4}{*}[-1.2em]{\shortstack{Ratio\\(\%)}}  & \multicolumn{6}{c}{Direct Competitors} & \multicolumn{5}{c}{Reference Results} 
\\
\cmidrule(lr){3-8} \cmidrule(lr){9-13} 
&
& \multicolumn{2}{c}{\shortstack{\textit{Coreset Selection}}} 
& \multicolumn{3}{c}{\shortstack{\textit{Decoupled Distill.}}}
& \multicolumn{1}{c}{\shortstack{\textit{Joint Distill.}}} & 
\multicolumn{5}{c}{\shortstack{\textit{Training on Full Text and/or Full Graphs}}}
\\
\cmidrule(lr){3-4} \cmidrule(lr){5-7} \cmidrule(lr){8-8} \cmidrule(lr){9-13} 
& 
& \multirow{2}{*}[-0.2em]{\scalebox{0.95}{Random}} 
& \multirow{2}{*}[-0.2em]{\scalebox{0.95}{K-Center}} 
& \multicolumn{4}{c}{\scalebox{0.95}{\shortstack{\textit{Distill. Text (\cmark) + Distill. Graph (\cmark)}}}} 
& \multicolumn{4}{c}{\scalebox{0.95}{\shortstack{\textit{Full Text (\xmark) + Distill. Graph (\cmark)}}}}
& \multirow{2}{*}[-0.2em]{\scalebox{0.95}{\shortstack{\textit{Full Text (\xmark) + } \\ \textit{Full Graph (\xmark)}}}} \\
\cmidrule(lr){5-8} \cmidrule(lr){9-12}
& 
& 
& 
& \scalebox{0.95}{TD} 
& \scalebox{0.95}{TDAL} 
& \scalebox{0.95}{CLM} 
& \multicolumn{1}{c}{\textbf{\scalebox{0.95}{\method\scalebox{0.8}{(Ours)}}}}
& \scalebox{0.95}{GCond} 
& \scalebox{0.95}{GCSNTK}\tablefootnote{The original code runs out of memory for several configurations in our setting. Therefore, we replace it with our own implementation instead (see Appendix~\ref{sec:app:implement:sntk}).}
& \scalebox{0.95}{GEOM} 
& \scalebox{0.95}{GCDM}
& \\
\midrule
\multirow[c]{3}{*}{\scalebox{0.95}{\shortstack{Cora\\(2.7K)}}}
& \scalebox{0.95}{2.5} 
& \accstd{62.1}{3.4} 
& \accstd{58.2}{3.7} 
& \accstd{79.3}{1.8} 
& \underline{\accstd{81.3}{0.7}}
& \accstd{80.0}{0.7} 
& \textbf{\accstd{84.4}{0.9}} 
& \accstd{84.1}{0.6} 
& \accstd{85.2}{0.7}
& \accstd{80.9}{0.9} 
& \accstd{78.6}{1.7} 
& \multirow[c]{3}{*}{\accstd{88.3}{0.5}} \\
& \scalebox{0.95}{5.0}
& \accstd{71.7}{3.9} 
& \accstd{70.4}{2.8} 
& \accstd{80.6}{1.9} 
& \underline{\accstd{83.8}{1.1}}
& \accstd{82.8}{0.6} 
& \textbf{\accstd{85.5}{1.3}} 
& \accstd{84.6}{0.8} 
& \accstd{85.4}{0.9}
& \accstd{83.8}{1.3} 
& \accstd{80.5}{1.2} 
& \\
& \scalebox{0.95}{7.5} 
& \accstd{79.6}{0.9} 
& \accstd{77.8}{0.4} 
& \accstd{81.1}{2.2} 
& \accstd{83.9}{0.6} 
& \underline{\accstd{84.1}{1.4}} 
& \textbf{\accstd{86.9}{1.1}} 
& \accstd{84.9}{0.9} 
& \accstd{85.8}{0.8}
& \accstd{84.6}{1.3} 
& \accstd{82.5}{1.0} 
& \\

\midrule

\multirow[c]{3}{*}{\scalebox{0.95}{\shortstack{Photo\\(48.4K)}}}
& \scalebox{0.95}{1.0} 
& \accstd{73.8}{3.5} 
& \accstd{71.5}{1.8} 
& \accstd{81.6}{0.4} 
& \underline{\accstd{82.5}{0.7}}
& \accstd{78.3}{0.9} 
& \textbf{\accstd{83.8}{0.4}} 
& \accstd{82.6}{0.2} 
& \accstd{83.5}{0.6}
& \accstd{82.5}{0.4} 
& \accstd{80.0}{0.8} 
& \multirow[c]{3}{*}{\accstd{86.6}{0.3}} \\
& \scalebox{0.95}{3.0} 
& \accstd{80.4}{1.4} 
& \accstd{74.2}{1.5} 
& \accstd{82.1}{0.7} 
& \underline{\accstd{83.6}{0.4}}
& \accstd{82.0}{0.6} 
& \textbf{\accstd{85.0}{0.4}} 
& \accstd{83.4}{0.6} 
& \accstd{83.4}{0.7} 
& \accstd{83.8}{0.4}
& \accstd{81.4}{0.6} 
& \\
& \scalebox{0.95}{5.0} 
& \accstd{81.4}{0.4} 
& \accstd{78.2}{0.7} 
& \accstd{82.6}{1.0} 
& \underline{\accstd{84.0}{0.3}}
& \accstd{83.0}{0.4} 
& \textbf{\accstd{85.8}{0.3}} 
& \accstd{83.6}{0.4} 
& \accstd{83.8}{0.9} 
& \accstd{84.1}{0.6}
& \accstd{81.5}{0.8} 
& \\

\midrule

\multirow[c]{3}{*}{\scalebox{0.95}{\shortstack{Computers\\(87.2K)}}}
& \scalebox{0.95}{1.0} 
& \accstd{78.9}{1.0} 
& \accstd{74.7}{1.4} 
& \accstd{83.9}{0.6} 
& \underline{\accstd{84.1}{0.6}}
& \accstd{82.1}{0.5} 
& \textbf{\accstd{86.4}{0.2}} 
& \accstd{84.6}{1.1} 
& \accstd{85.6}{0.3}
& \accstd{85.2}{0.8} 
& \accstd{82.6}{0.3} 
& \multirow[c]{3}{*}{\accstd{90.1}{0.3}} \\
& \scalebox{0.95}{2.0} 
& \accstd{81.0}{0.6} 
& \accstd{75.3}{0.9} 
& \accstd{85.3}{0.6} 
& \underline{\accstd{85.3}{0.3}}
& \accstd{84.2}{0.3} 
& \textbf{\accstd{87.4}{0.4}} 
& \accstd{84.8}{0.3} 
& \accstd{86.3}{0.4}
& \accstd{86.1}{0.9} 
& \accstd{83.8}{0.3} 
& \\
& \scalebox{0.95}{3.0} 
& \accstd{83.3}{0.4} 
& \accstd{79.2}{0.8} 
& \accstd{85.8}{0.2} 
& \underline{\accstd{86.0}{0.4}}
& \accstd{84.9}{0.3} 
& \textbf{\accstd{87.9}{0.4}} 
& \accstd{84.9}{0.4} 
& \accstd{86.4}{0.5} 
& \accstd{86.4}{0.9}
& \accstd{84.2}{0.5} 
& \\

\midrule

\multirow[c]{3}{*}{\scalebox{0.95}{\shortstack{Arxiv\\(169.3K)}}}
& \scalebox{0.95}{0.5} 
& \accstd{64.7}{0.3} 
& \accstd{59.2}{0.6} 
& \accstd{66.9}{1.2} 
& \underline{\accstd{68.2}{0.3}}
& \accstd{68.1}{0.2} 
& \textbf{\accstd{72.3}{0.1}} 
& \accstd{70.1}{0.3} 
& \accstd{69.6}{0.3} 
& \accstd{71.7}{0.5}
& \accstd{69.3}{0.3} 
& \multirow[c]{3}{*}{\accstd{75.7}{0.2}} \\
& \scalebox{0.95}{1.0} 
& \accstd{66.9}{0.4} 
& \accstd{61.0}{0.3} 
& \accstd{68.4}{0.5} 
& \accstd{69.4}{0.4} 
& \underline{\accstd{69.7}{0.3}}
& \textbf{\accstd{73.3}{0.3}} 
& \accstd{70.3}{0.4} 
& \accstd{70.2}{0.5} 
& \accstd{72.2}{0.6}
& \accstd{69.5}{0.3} 
& \\
& \scalebox{0.95}{1.5} 
& \accstd{68.8}{0.6} 
& \accstd{63.3}{0.8} 
& \accstd{69.7}{0.4} 
& \underline{\accstd{70.6}{0.4}}
& \accstd{70.3}{0.3} 
& \textbf{\accstd{74.0}{0.2}} 
& \accstd{70.3}{0.3} 
& \accstd{70.9}{0.2} 
& \accstd{72.3}{0.7}
& \accstd{69.5}{0.2} 
& \\

\bottomrule
\end{tabular}%
};

\draw[line width=0.3pt]
  ($(tbl.north west)+(9.28cm,-0.02cm)$) --
  ($(tbl.south west)+(9.28cm, 0.02cm)$);

\draw[line width=0.3pt]
  ($(tbl.north west)+(9.32cm,-0.02cm)$) --
  ($(tbl.south west)+(9.32cm, 0.02cm)$);
\end{tikzpicture}
\caption{Performance (test accuracy in \%) of dataset distillation methods on semi-supervised node classification. Results to the left of the vertical lines correspond to the direct competitors, while those to the right are provided as reference results. The ratio denotes $\frac{M}{N}$, i.e., the relative size of the synthetic dataset. 
Note that \method enables \textit{joint distillation} of text and graph data due to its efficiency. In contrast, the baselines adopt \textit{decoupled distillation}, where each text distillation method is paired with the best-performing graph distillation method.
The best and second-best results among the direct competitors are highlighted in \textbf{bold} and \underline{underlined}, respectively.}
\label{tab:main_results}
\end{table*}

\subsection{Experiment Settings}
\label{sec:exp:settings}

\paragraph{Dataset.}
We evaluate \method on four representative TAG datasets: two academic network datasets, Cora~\cite{sen2008collective} and Arxiv~\cite{hu2020open}, and two e-commerce network datasets, Photo and Computers~\cite{shchur2018pitfalls}. Dataset details are provided in Appendix~\ref{sec:app:dataset}.

\paragraph{Baseline.}
We consider two competitor groups: \textit{coreset selection} and \textit{decoupled distillation}. Coreset selection, which includes Random and K-Center~\cite{sener2017active}, simply samples representative subsets from the full training dataset. Decoupled distillation combines existing text and graph dataset distillation methods under a decoupled training pipeline, distilling text and graph separately. For text distillation, we use TD~\cite{li2021data}, TDAL~\cite{maekawa2023dataset}, and CondenseLM (CLM)~\cite{shen2025condenselm}. For graph distillation, we use GCond~\cite{jin2021graph}, GCSNTK~\cite{wang2024fast}, GEOM~\cite{zhang2024navigating}, and GCDM~\cite{liu2022graph}.

In addition, for references, we report results under \textit{training on full text and/or full graphs} using the decoupled training pipeline.
Note that they are not direct competitors of \method, as they require access to full text and/or graph data, deviating from the objective of standard dataset distillation.  
Refer to Appendix~\ref{sec:app:implement:baseline} for more baseline details.

\paragraph{Implementation.}
We use BERT$_{\text{base}}$~\cite{devlin2019bert} with LoRA~\cite{hu2022lora} as the pretrained LM and a two-layer GCN~\cite{kipf2016semi} as the GNN, which are widely adopted backbone choices in prior studies. Implementation details and hyperparameter settings are provided in Appendix~\ref{sec:app:implement_details}.

\subsection{Main Results}
\paragraph{Performance comparison.}\label{sec:exp:main} Following the common protocol, we evaluate each dataset distillation method by training an LM--GNN model\tablefootnote{Refer to Section~\ref{sec:exp:settings} for model details. The superiority of our method is robust to model choices. See the cross-architecture results on GNNs and also Appendix~\ref{sec:app:backbone} for results with varying LMs.}
using its output synthetic dataset and testing it on the full dataset.
Both training and test are conducted on semi-supervised node classification (Section~\ref{sec:prelim:concepts}).
The results are presented in Table~\ref{tab:main_results}.

First, \method consistently outperforms all decoupled distillation baselines, which distill text and graph separately, as well as coreset selection baselines, across all experimental settings. This highlights the effectiveness of our key idea, i.e., jointly distilling both modalities in an end-to-end manner.

Second, \method even outperforms four graph-only distillation methods, which rely on the full text for training, in 11 of 12 cases. This suggests that the performance gains are not simply due to stronger LM representations, but are more likely to come from jointly distilling text and graph information, which is the strength of our method.

Lastly, \method often achieves performance close to that of full-dataset (i.e., both full graph and text) training. For example, on Arxiv, it preserves 97\% of the full-dataset performance using a synthetic dataset of size 1\% of the full dataset.
On Photo, it retains 99\% of the full-dataset performance using a synthetic dataset of size 5\%.

\begin{table}[t]
\centering
    \small
    \setlength{\tabcolsep}{1.5pt}
    \begin{tabular}{ccccccc}
    \toprule
    Dataset & Method & GCN & SGC & SAGE & APPNP & Avg. \\
    \midrule
    
    \multirow{4}{*}{\shortstack{Computers\\(2\%)}} 
    & GCSNTK  & 86.3 & 85.8 & 85.6 & 85.1 & 85.7 \\
    & GEOM    & 86.1 & 84.9 & 85.3 & 85.2 & 85.4 \\
    & TDAL+GEOM   & 85.3 & 82.7 & 83.0 & 82.9 & 83.5 \\
    & \method & \textbf{87.4} & \textbf{86.1} & \textbf{86.8} & \textbf{85.8} & \textbf{86.5} \\
    \midrule
    
    \multirow{4}{*}{\shortstack{Arxiv\\(1\%)}} 
    & GCSNTK  & 70.2 & 70.4 & 71.1 & 70.4 & 70.5 \\
    & GEOM    & 72.2 & 71.2 & 72.5 & \textbf{72.8} & 72.2 \\
    & TDAL+GEOM    & 69.4 & 68.2 & 68.7 & 69.7 & 69.0 \\
    & \method & \textbf{73.3} & \textbf{73.0} & \textbf{73.3} & 72.5 & \textbf{73.0} \\
    \bottomrule
    \end{tabular}
\caption{\label{tab:cross_arch_gnn}Cross-architecture generalization of dataset distillation methods, with best results in \textbf{bold}.} 
\end{table}

\paragraph{Cross-architecture generalization.}
We further evaluate the output synthetic datasets across different GNN architectures. In this experiment, the datasets are distilled using GCN (see Section~\ref{sec:exp:settings} for details), and the resulting synthetic data are used to train models with multiple GNN architectures, including SGC~\cite{chen2020simple}, GraphSAGE~\cite{hamilton2017inductive} and APPNP~\cite{gasteiger2018predict}. 

As shown in Table~\ref{tab:cross_arch_gnn}, \method achieves the best performance for most architectures.
These results indicate that the resulting synthetic datasets do not overfit to the specific GNN architecture used during distillation and generalize well across architectures.

\paragraph{Extra experiments.} See Appendix~\ref{sec:app:backbone}--\ref{sec:app:large} for extra experiments with different LM backbones, under inductive settings, and on large-scale datasets.

\begin{figure}
    \centering
        \includegraphics[width=\linewidth]{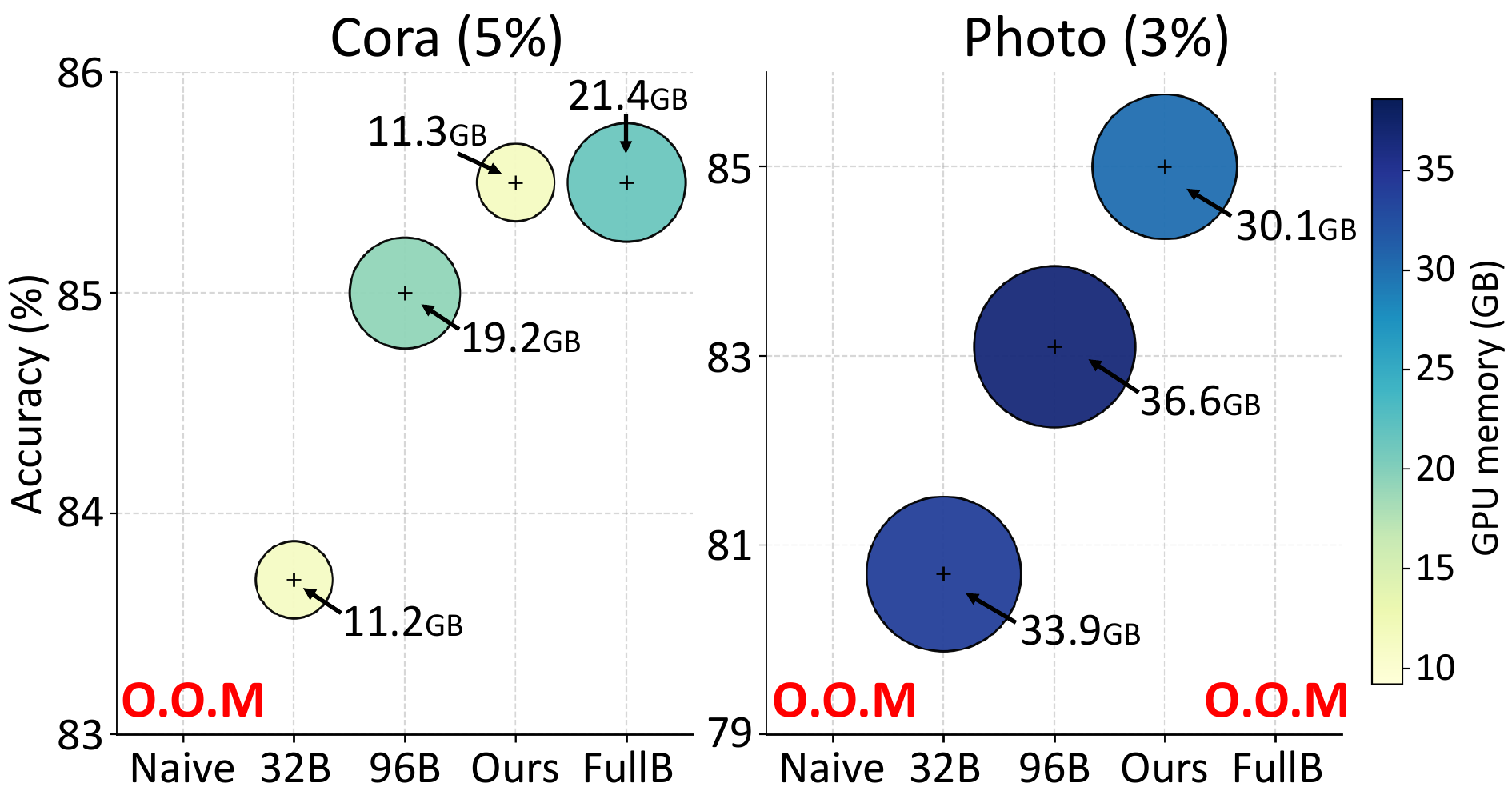}
        \caption{\label{fig:ablation}Ablation study. Circle size indicates GPU memory usage. \texttt{Naive} denotes the variant that uses a trainable GNN for the outer-loop optimization instead of our kernel-based formulation. \texttt{FullB} denotes full-batch training without batch-wise gradient injection, and \texttt{32B}/\texttt{96B} denote naive mini-batch training with batch sizes of 32 and 96, respectively. Ours employs batch-wise gradient injection with a batch size of 64.}
\end{figure}

\subsection{Ablation Study}
We analyze how the efficiency and effectiveness of \method are affected by its key design components: (1) using a graph-aware neural tangent kernel for outer-loop optimization (see Section~\ref{sec:method:outer}) and (2) batch-wise gradient injection (see Section~\ref{sec:method:bgi}).
The results are presented in Figure~\ref{fig:ablation}.

\paragraph{Outer-loop optimization.}
We compare against a naive variant that replaces our kernel-based formulation with the same trainable GNN used in the inner-loop. This naive variant already runs out of memory even on the smallest dataset, Cora, whereas \method remains feasible across all considered datasets, including Cora and Photo.

\paragraph{Batch-wise gradient injection.} We compare \method against full-batch training without batch-wise gradient injection, as well as naive mini-batch training with batch sizes of 32 and 96. On Cora, \method, which employs batch-wise gradient injection with a batch size of 64, matches the full-batch performance while outperforming the mini-batch variants. On Photo, where full-batch training runs out of memory, \method remains feasible and outperforms all mini-batch alternatives. Notably, it substantially outperforms the variant with a batch size of 32 while using even less GPU memory. 

These results show that both components are essential to \method: the graph-aware neural tangent kernel enables efficient outer-loop optimization, while batch-wise gradient injection preserves the dataset distillation objective under mini-batch training and improves both efficiency and performance.

See Appendix~\ref{sec:app:ablationa_kernel}--\ref{sec:app:ablation_GNN} for ablation studies on the graph-aware kernel and the inner-loop GNN.

\begin{figure}[t]
    \centering
    \hspace*{0.06\linewidth}\includegraphics[width=0.8\linewidth]{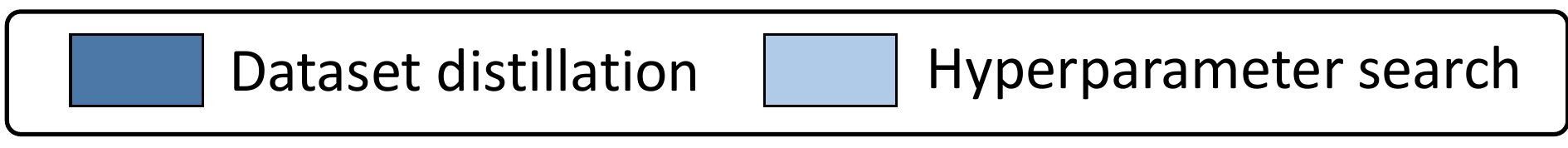}
    \vspace{0.2em}

    \begin{subfigure}{0.49\columnwidth}
        \centering
        \includegraphics[width=\linewidth]{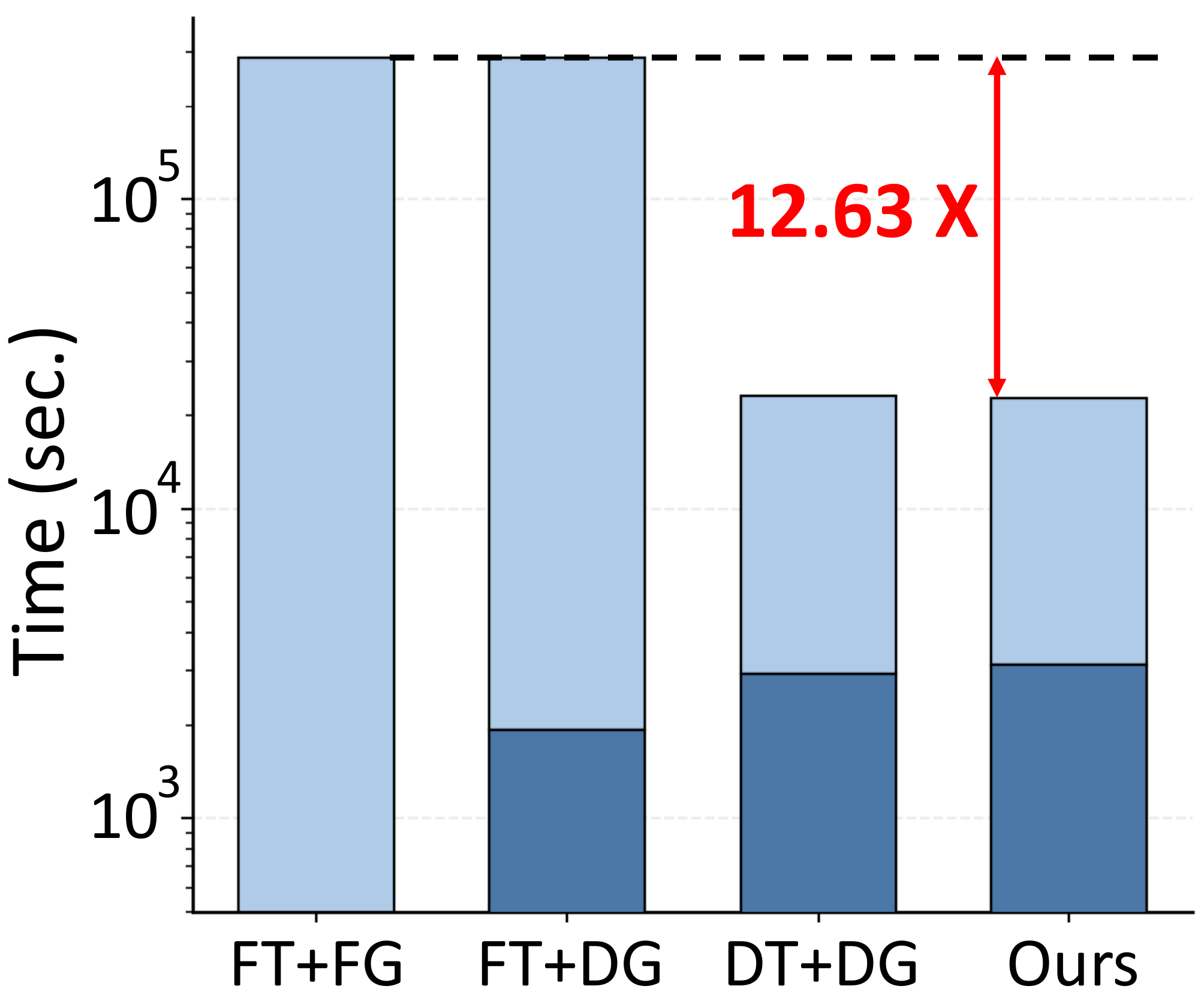}
        \caption{Photo}
        \label{fig:search_photo}
    \end{subfigure}\hfill
    \begin{subfigure}{0.49\columnwidth}
        \centering
        \includegraphics[width=\linewidth]{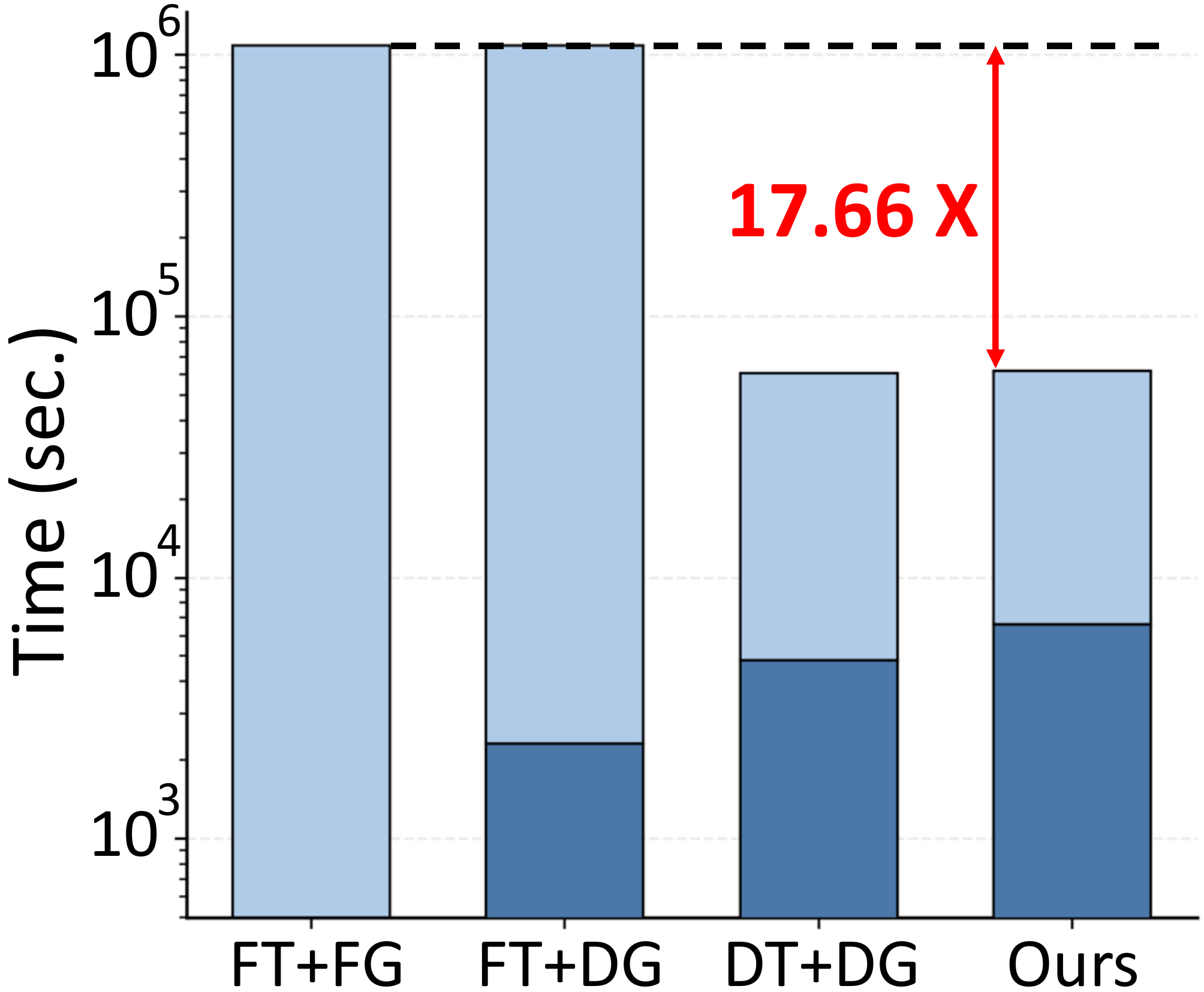}
        \caption{Arxiv}
        \label{fig:search_arxiv}
    \end{subfigure}
    \caption{\label{fig:search}Hyperparameter search time under four settings: full text and full graph (\texttt{FT+FG}), full text and distilled graph (\texttt{FT+DG}), separately distilled text and graph (\texttt{DT+DG}), and jointly distilled text and graph (ours).}
\end{figure}

\subsection{Application: Hyperparameter Search}
We evaluate dataset distillation methods by applying them to hyperparameter search.
Synthetic datasets produced by dataset distillation can accelerate hyperparameter search by reducing the time required for repeated training.
Specifically, we consider LM hyperparameter search over 192 configurations, including batch size, learning rate, rank, scaling factor, and LoRA dropout. 
We compare full text and full graph (\texttt{FT+FG}), full text and distilled graph (\texttt{FT+DG}), separately distilled text and graph (\texttt{DT+DG}), and jointly distilled text and graph (ours). We use TDAL and GEOM for text and graph dataset distillation, respectively.
The results are reported in Figure~\ref{fig:search}, and
reported time includes, where applicable, the time required for dataset distillation, which is a one-time cost (see Appendix~\ref{sec:app:cost} for detailed results).

Because LM training dominates the cost in TAG learning, methods that use the full text, namely \texttt{FT+FG} and \texttt{FT+DG}, require much longer search time. Even after accounting for dataset distillation time, ours achieves 12.63$\times$ and 17.66$\times$ speed-ups over hyperparameter search on the full dataset (\texttt{FT+FG}) on Photo and Arxiv. 
Ours remains competitive with \texttt{DT+DG} in terms of overall search time.

\section{Conclusion}
\label{sec:conc}
In this work, we present \method, the first dataset distillation method for TAGs that enables efficient joint distillation of text and graph data under joint LM--GNN training, without decoupling.
To achieve this, \method integrates an LM with a graph-aware neural tangent kernel and introduces batch-wise gradient injection.
Our experiments show that \method consistently outperforms baselines across various scenarios, suggesting dataset distillation as a viable direction for efficient TAG learning.

\section*{Limitations}
\label{sec:limit}
While we propose an effective method for text-attributed graph dataset distillation, several limitations remain. First, the current method performs distillation at the token-embedding level and thus does not directly produce human-interpretable synthetic data in raw-text form. Extending the framework toward more interpretable text-level distillation would be a valuable direction for future work.
Second, our current formulation is developed for semi-supervised node classification. Although this is a primary downstream task in text-attributed graph learning, extending the method to self-supervised settings and broader downstream tasks would be an important direction for future research.
Third, while our experiments show promising results across multiple benchmarks, further evaluation on million-scale text-attributed graphs would help better understand the scalability and generality of the proposed framework.

\section*{Ethical Considerations}
\label{sec:ethical}
We comply with the ACL Ethics Policy\footnote{\url{https://www.aclweb.org/portal/content/acl-code-ethics}} in this work. All experiments are conducted on publicly available benchmark datasets that have been widely used in prior research. This work does not involve collecting new personal data and uses the datasets in their publicly released form, without additional preprocessing or annotation.

\section*{Acknowledgements}
This work was supported by Institute of Information \& communications Technology Planning \& Evaluation (IITP) grant funded by the Korea government (MSIT) (No. RS-2024-00438638, EntireDB2AI: Foundations and Software for Comprehensive Deep Representation Learning and Prediction on Entire Relational Databases, 40\%) (No. RS-2024-00457882, National AI Research Lab Project, 50\%) (RS-2019-II190075, Artificial Intelligence Graduate School Program (KAIST), 10\%).


\bibliography{reference}

\appendix
\label{sec:appendix}
\section*{Appendix}
\label{sec:app}
\section{Dataset Details}
\label{sec:app:dataset}

\begin{table}[!htbp]
    \centering
    \begin{tabular}{lrrr}
        \toprule
        Dataset & $|V|$ & $|E|$ & \# Class \\
        \midrule
        Cora      & 2,708   & 5,429     & 7  \\
        Photo     & 48,362  & 500,928   & 12 \\
        Computers & 87,229  & 721,081   & 10 \\
        Arxiv     & 169,343 & 1,166,243 & 40 \\
        \bottomrule
    \end{tabular}
    \caption{Dataset statistics. $|V|$ and $|E|$ denote the numbers of nodes and edges, respectively.}
    \label{tab:dataset_stats}
\end{table}

\noindent
In this section, we describe the datasets used in our experiments. We use the raw text for each dataset provided by \citet{he2023harnessing} and \citet{yan2023comprehensive} and adopt a 60/20/20 split for the training, validation, and test sets. Dataset statistics are summarized in Table~\ref{tab:dataset_stats}.

\paragraph{Cora.}
Cora~\cite{sen2008collective} is a citation network dataset in which nodes represent scientific papers and edges denote citation links between papers. The text attribute of each node consists of the paper title and abstract, and the node label indicates the research topic.

\paragraph{Photo.}
Photo~\cite{shchur2018pitfalls} is an Amazon co-purchase network dataset in the photo-related product category, where nodes represent products and edges indicate that two products are frequently purchased together. The text attribute of each node consists of product descriptions and reviews, and the node label indicates the product category.

\paragraph{Computers.}
Computers~\cite{shchur2018pitfalls} is an Amazon co-purchase network dataset in the computer-related product category, where nodes represent products and edges indicate that two products are frequently purchased together. The text attribute of each node consists of product descriptions and reviews, and the node label indicates the product category.

\paragraph{Arxiv.}
Arxiv~\cite{hu2020open} is a citation network dataset of arXiv papers, where nodes represent papers and edges denote citation relationships. The text attribute of each node consists of the paper title and abstract, and the node label indicates the subject area of the paper.

\begin{table*}[t]
    \centering
    \begin{tabular}{lcccccccc}
        \toprule
        Dataset & $K$ & $P_{\text{in}}$ & $T_{\text{in}}$ & $P_{\text{out}}$ & $T_{\text{out}}$ & $\tilde{\eta}$ & $\lambda$ & Batch size\\
        \midrule
        Cora      & 80 & 10 & 90 & 10 & 8 & $5\times10^{-4}$ & $1\times10^{-4}$ & 32 \\
        Photo     & 50 & 10 & 60 & 10 & 8 & $1\times10^{-4}$ & $1\times10^{-5}$ & 64\\
        Computers & 50 & 8 & 80 & 8 & 8 & $1\times10^{-4}$ & $1\times10^{-5}$ & 64\\
        Arxiv     & 50 & 6 & 60 & 8 & 5 & $1\times10^{-4}$ & $1\times10^{-5}$ & 64\\
        \bottomrule
    \end{tabular}
    \caption{Hyperparameter settings for distillation. $P_{\text{in}}$ and $P_{\text{out}}$ denote the early stopping patience for the inner- and the outer-loop, respectively. $K$ denotes the number of training epochs, $T_{\text{in}}$ and $T_{\text{out}}$ denote the inner- and outer-loop iteration steps, respectively, and $\lambda$ is the ridge coefficient.}
    \label{tab:distill_hyper}
\end{table*}

\section{Experiment Details}
\label{sec:app:implement}

\subsection{Baselines}
\label{sec:app:implement:baseline}

In this section, we provide additional implementation details for the baseline experiments. For \textit{Coreset Selection} methods, we first select a subset of the original dataset according to each method and then train the model using the joint training pipeline in Eq.~\eqref{eq:joint}. For \textit{Decoupled Distill.} baselines, we distill the text and graph components separately according to each baseline method, and then train the final model on the resulting distilled text and distilled graph under the decoupled training pipeline in Eq.~\eqref{eq:stage1_decoupled}--\eqref{eq:stage2_decoupled}. For \textit{Full Text + Distill. Graph} baselines, we first train the LM on the full text data, use the resulting text representations as the initial input features for the GNN, and distill only the graph component. The final model is then trained on the full text and the distilled graph under the same decoupled training pipeline. Below, we describe only the baselines whose settings require further clarification.

\paragraph{GCond/GCDM.}
Both methods propose two variants for the synthetic graph structure: one that learns the synthetic graph structure and another that uses an identity matrix. We report the latter for the baselines, as it performs better in our setting.

\paragraph{CondenseLM.}
As the official code was not publicly available during our experiments, we use our own implementation based on the description in the paper.
For the LLM component, we use GPT-4o-mini.

\subsection{Implementation Details}
\label{sec:app:implement_details}
For each method, we repeat distillation 5 times, and repeat training and evaluation 5 times for each distilled dataset. All experiments were conducted on two NVIDIA RTX PRO 6000 GPUs, each with 96GB of memory.

For hyperparameter settings, we tune the outer-loop learning rate $\tilde{\eta}$ from $\{1\times10^{-3}, 5\times10^{-4}, 1\times10^{-4}, 5\times10^{-5}, 1\times10^{-5}\}$ and the ridge coefficient $\lambda$ from $\{1\times10^{-3}, 1\times10^{-4}, 1\times10^{-5}\}$. We use the same LM and GNN training hyperparameters for both our method and the baselines. For LM training, we use a learning rate $\eta=1\times10^{-4}$, a maximum sequence length $Q=256$, and apply LoRA with $r=4$, $\alpha=8$, and dropout rate 0.05. For GNN training, we use a two-layer model with hidden dimension 256 and dropout rate 0.3. We use the Adam optimizer~\cite{kingma2014adam} for all optimization procedures. The selected hyperparameter values for each dataset are reported in Table~\ref{tab:distill_hyper}.

\subsection{Details of Hyperparameter Search}
\label{sec:app:cost}

We provide a detailed running time breakdown in Table~\ref{tab:cost_breakdown}, separating one-time distillation cost from hyperparameter search cost. \textit{Full Text + Distill. Graph} requires less distillation time because only the graph is distilled, but its hyperparameter search time is much larger since searching for LM hyperparameters still requires training on the full text, which becomes the main bottleneck. Compared with \textit{Distill. Text + Distill. Graph}, \method has a slightly higher one-time distillation cost because it jointly distills text and graph, while its total search time remains comparable.

\begin{table}[t]
\centering
\small
\setlength{\tabcolsep}{2.5pt}
\begin{tabular}{@{}ccccc@{}}
\toprule
Dataset & Time & \shortstack{\textit{Full Text} +\\\textit{Distill. Graph}} & \shortstack{\textit{Distill. Text} +\\\textit{Distill. Graph}} & \method \\
\midrule
\multirow{3}{*}{Photo}
& Distillation & 1.9K & 2.9K & 3.1K \\
& Search & 285.4K & 20.2K & 19.6K \\
& Total & 287.3K & 23.1K & 22.7K \\
\midrule
\multirow{3}{*}{Arxiv}
& Distillation & 2.3K & 4.8K & 6.6K \\
& Search & 1.09M & 55.6K & 54.9K \\
& Total & 1.09M & 60.4K & 61.5K \\
\bottomrule
\end{tabular}
\caption{Running time breakdown of \method and baselines. All values are reported in seconds. Distillation denotes one-time distillation cost, and Search denotes hyperparameter search cost.}
\label{tab:cost_breakdown}
\end{table}

\subsection{Memory-efficient Implementation of SNTK}
\label{sec:app:implement:sntk}

Recall from Eq.~\eqref{eq:LMrep}--\eqref{eq:kerenl_form2} that \method computes node representations
$\mathbf{H}_{\mathcal{T}} = f_{\phi}(\mathbf{X})$ and $\mathbf{H}_{\mathcal{S}} = f_{\phi}(\mathbf{X}')$, and then forms the kernel matrices
$\mathbf{K}_{\mathcal{T}\mathcal{S}}$ and $\mathbf{K}_{\mathcal{S}\mathcal{S}}$ using SNTK.
In our setting, SNTK is applied either between $(\mathbf{H}_{\mathcal{T}}, \mathbf{A})$ and $(\mathbf{H}_{\mathcal{S}}, \mathbf{I})$
to construct $\mathbf{K}_{\mathcal{T}\mathcal{S}}$, or between $(\mathbf{H}_{\mathcal{S}}, \mathbf{I})$ and $(\mathbf{H}_{\mathcal{S}}, \mathbf{I})$
to construct $\mathbf{K}_{\mathcal{S}\mathcal{S}}$.
Below, we describe how $\mathbf{K}_{\mathcal{T}\mathcal{S}}$ is computed in our implementation.
The computation of $\mathbf{K}_{\mathcal{S}\mathcal{S}}$ is analogous, with $\mathbf{I}$ on both sides.

\paragraph{Overview of SNTK.}
We first initialize
\[
\Sigma_{\mathcal{T}\mathcal{S}} = \Theta_{\mathcal{T}\mathcal{S}} = \mathbf{H}_{\mathcal{T}} \mathbf{H}_{\mathcal{S}}^\top.
\]
Here, $\Sigma_{\mathcal{T}\mathcal{S}}$ denotes the intermediate kernel block, while
$\Theta_{\mathcal{T}\mathcal{S}}$ denotes the neural tangent kernel (NTK) block.
Both matrices are recursively updated through neighborhood aggregation and nonlinear kernel transformation, and the final $\Theta_{\mathcal{T}\mathcal{S}}$ is used as $\mathbf{K}_{\mathcal{T}\mathcal{S}}$.

\paragraph{Efficient neighborhood aggregation.}
The original SNTK implementation performs neighborhood aggregation by constructing a sparse Kronecker-product matrix and applying it to the vectorized kernel matrix.
This formulation expresses aggregation over all node pairs as a single linear operation on the cross-kernel.
However, because the operator is defined on the Cartesian product of the two node sets, its size grows with the product of their numbers of nodes, incurring substantial memory overhead.

In our setting, the synthetic graph is fixed as $\mathbf{A}'=\mathbf{I}$.
Therefore, neighborhood aggregation on the synthetic side becomes the identity operation, and the Kronecker-based update reduces to applying the graph aggregation operator only on the full-graph side. Accordingly, the aggregation update becomes
\[
\Sigma_{\mathcal{T}\mathcal{S}} \leftarrow \mathbf{A}\Sigma_{\mathcal{T}\mathcal{S}},
\qquad
\Theta_{\mathcal{T}\mathcal{S}} \leftarrow \mathbf{A}\Theta_{\mathcal{T}\mathcal{S}}.
\]
Equivalently, with $\mathrm{vec}(\cdot)$ denoting vectorization and $\otimes$ the Kronecker product, we have
\[
\mathrm{vec}(\mathbf{A}\Sigma_{\mathcal{T}\mathcal{S}})
=
(\mathbf{I}\otimes \mathbf{A})\,\mathrm{vec}(\Sigma_{\mathcal{T}\mathcal{S}}).
\]
The same identity also holds when $\mathbf{A}$ is replaced by its degree-normalized form $\mathbf{D}^{-1}\mathbf{A}$, where $\mathbf{D}$ is the degree matrix of $\mathbf{A}$.
Thus, our implementation computes exactly the same neighborhood aggregation as the original Kronecker-based formulation, while avoiding explicit construction of the large Kronecker-product matrix.

\begin{table}[t]
    \centering
    \small
    \setlength{\tabcolsep}{2.5pt}
    \begin{tabular}{cccccc}
        \toprule
        Dataset & LM & GCSNTK & GEOM & TDAL & \method \\
        \midrule
        \multirow[c]{2}{*}{\shortstack{Cora \\ (5\%)}} 
            & RoBERTa & 84.0 & 82.4 & 80.5 & \textbf{85.5} \\
            & DeBERTa & 84.6 & 82.3 & 82.4 & \textbf{86.0} \\
        \midrule
        \multirow[c]{2}{*}{\shortstack{Photo \\ (3\%)}} 
            & RoBERTa & 82.9 & 83.5 & 82.7 & \textbf{85.2} \\
            & DeBERTa & 82.4 & 83.6 & 81.2 & \textbf{84.9} \\
        \midrule
        \multirow[c]{2}{*}{\shortstack{Computers \\ (2\%)}} 
            & RoBERTa & 86.6 & 85.2 & 86.4 & \textbf{87.8} \\
            & DeBERTa & 85.7 & 86.4 & 85.6 & \textbf{87.2} \\
        \midrule
        \multirow[c]{2}{*}{\shortstack{Arxiv \\ (1\%)}} 
            & RoBERTa & 70.5 & 72.5 & 69.9 & \textbf{72.8} \\
            & DeBERTa & 70.5 & 72.4 & 70.9 & \textbf{73.4} \\
        \bottomrule
    \end{tabular}
    \caption{Performance comparison across different LM backbones. We report the performance of TDAL combined with the best-performing graph dataset distillation method for each configuration. The best results are highlighted in \textbf{bold}.}
    \label{tab:backbone}
\end{table}
 
\begin{table}[t]
\centering
\small
\setlength{\tabcolsep}{3pt}
\begin{tabular}{@{}ccccc@{}}
\toprule
Dataset & Ratio (\%) & \shortstack{TDAL+\\GCSNTK} & \shortstack{TDAL+\\GEOM} & \method \\
\midrule
Arxiv-ind & 0.5 & 66.1 & 64.6 & \textbf{71.4} \\
\bottomrule
\end{tabular}
\caption{Node classification performance under the inductive Arxiv setting. The best result is highlighted in \textbf{bold}.}
\label{tab:inductive}
\end{table}

\paragraph{Efficient approximation for normalization.}
In SNTK, each recursive update of the cross-kernel requires normalization terms from the corresponding self-kernels, namely
\[
\mathbf{p}_{\mathcal{T}}^{(k)}=\sqrt{\mathrm{diag}\!\left(\Sigma_{\mathcal T\mathcal T}^{(k)}\right)},
\qquad
\mathbf{p}_{\mathcal{S}}^{(k)}=\sqrt{\mathrm{diag}\!\left(\Sigma_{\mathcal S\mathcal S}^{(k)}\right)}.
\]
In the original implementation, these quantities are obtained by explicitly maintaining the full self-kernel matrices throughout the recursion and reading their diagonals when needed. This is memory-inefficient, since only the diagonal values are used for normalization.

To avoid this overhead, we use a feature-based approximation of the normalization vectors. At aggregation round $k$, we propagate
\[
\mathbf{H}_{\mathcal{T}}^{(k)}=\mathbf{A}\mathbf{H}_{\mathcal{T}}^{(k-1)},
\]
or, when $\mathbf{A}$ is replaced by its degree-normalized form $\mathbf{D}^{-1}\mathbf{A}$, as
\[
\mathbf{H}_{\mathcal{T}}^{(k)}=\mathbf{D}^{-1}\mathbf{A}\mathbf{H}_{\mathcal{T}}^{(k-1)},
\]
and define the approximate normalization vector by the row-wise $\ell_2$ norms:
\[
\tilde{\mathbf{p}}_{\mathcal{T}}^{(k)}
=
\big[
\|\mathbf{H}_{\mathcal{T},1}^{(k)}\|_2,\,
\ldots,\,
\|\mathbf{H}_{\mathcal{T},N}^{(k)}\|_2
\big]^\top.
\]
The same construction is applied to the synthetic side, where the synthetic graph is fixed as $\mathbf{I}$ and the propagated representations therefore remain unchanged across aggregation rounds. This construction is exact at the first aggregation round. When multiple aggregation rounds are used, later rounds rely on a memory-efficient feature-based approximation of the normalization terms.

\section{Additional Experiments}\label{sec:app:addtional}

\subsection{Generalization across LM Backbones}\label{sec:app:backbone}
We further evaluate our dataset distillation method with different LM backbones, RoBERTa$_{\text{base}}$~\cite{liu2019roberta} and DeBERTa$_{\text{base}}$~\cite{he2021deberta}. As shown in Table~\ref{tab:backbone}, \method consistently outperforms all baselines across all datasets for both backbone choices.

\subsection{Experiment under an Inductive Setting}\label{sec:app:inductive}

To evaluate \method in an inductive setting, we convert Arxiv into an inductive graph dataset, denoted as Arxiv-ind. We construct timestamp-based train/validation/test splits with a 60:20:20 ratio, where validation and test nodes are not observed during the training period. We compare \method with strong decoupled distillation baselines, including TDAL+GCSNTK and TDAL+GEOM. As shown in Table~\ref{tab:inductive}, \method continues to outperform the decoupled baselines under this inductive setting.

\begin{table}[t]
    \centering
    \small
    \setlength{\tabcolsep}{3pt}
    \begin{tabular}{@{}ccccc@{}}
    \toprule
    Dataset & Ratio (\%) & \shortstack{TDAL+\\GCSNTK} & \shortstack{TDAL+\\GEOM} & \method \\
    \midrule
    Goodreads-NC & 0.1 & 76.8 & 75.6 & \textbf{78.7} \\
    \bottomrule
    \end{tabular}
    \caption{Node classification performance on Goodreads-NC. The best result is highlighted in \textbf{bold}.}
    \label{tab:goodread}
\end{table}

\subsection{Experiment on a Large-scale Dataset}\label{sec:app:large}

To further evaluate \method on a larger graph, we conduct an additional experiment on Goodreads-NC~\cite{zhu2025mosaic}, which is substantially larger than the datasets used in the main experiments, containing 685K nodes and 7.2M edges. Although Goodreads-NC is a multimodal-attributed graph dataset, we use only the text attributes in this experiment to match the TAG setting considered in this paper. As shown in Table~\ref{tab:goodread}, \method continues to outperform the decoupled distillation baselines on this larger dataset.

\begin{table}[t]
\centering
    \small
    \setlength{\tabcolsep}{3.5pt}
    \begin{tabular}{ccccc}
    \toprule
    Dataset & Ratio (\%) & Dot Product & NTK & Ours \\
    \midrule
    Cora & 2.5 & 78.8 & 82.8 & \textbf{84.4} \\
    Photo & 1.0 & 76.5 & 80.5 & \textbf{83.8} \\
    Computers & 1.0 & 76.6 & 84.9 & \textbf{86.4} \\
    \bottomrule
    \end{tabular}
    \caption{Ablation study on the kernel used for the outer-loop optimization, with best results in \textbf{bold}.}
    \label{tab:kernel_ablation}
\end{table}

\subsection{Ablation on the Graph-aware Kernel}\label{sec:app:ablationa_kernel}

To examine the effect of our graph-aware kernel built on LM representations, we compare it with structure-agnostic alternatives: a dot-product kernel and a standard NTK~\cite{jacot2018neural}. As shown in Table~\ref{tab:kernel_ablation}, the graph-aware kernel performs best, suggesting that incorporating graph structure into the outer-loop KRR objective is crucial for effective distillation.

\begin{table}[t]
    \centering
    \small
    \setlength{\tabcolsep}{2pt}
    \begin{tabular}{@{}ccccc@{}}
    \toprule
    \textit{Distill.} & \textit{Final} & Cora (2.5\%) & Photo (1\%) & Computers (1\%) \\
    \midrule
    GCN & GCN & \textbf{84.4} & \textbf{83.8} & \textbf{86.4} \\
    MLP & GCN & 83.8 & 83.5 & 86.1 \\
    \midrule
    SAGE & SAGE & \textbf{82.2} & \textbf{84.0} & \textbf{85.5} \\
    MLP & SAGE & 80.0 & 83.2 & 84.4 \\
    \midrule
    GIN & GIN & \textbf{83.4} & \textbf{83.8} & \textbf{83.6} \\
    MLP & GIN & 82.1 & 82.2 & 82.8 \\
    \bottomrule
    \end{tabular}
\caption{Ablation study on the inner-loop GNN. \textit{Distill.} denotes the model used during distillation, and \textit{Final} denotes the GNN architecture of the final LM--GNN model trained on the distilled dataset. The best results within each group are highlighted in \textbf{bold}.}
\label{tab:innerloop}
\end{table}

\subsection{Ablation on the Inner-loop GNN}\label{sec:app:ablation_GNN}

Although $\mathbf{A}'=\mathbf{I}$ limits nontrivial message passing among synthetic nodes, we use a GNN module on top of LM representations in the inner-loop to match the integrated LM--GNN model used in final training. To examine the effectiveness of this design choice, we evaluate replacing the inner-loop GNN with an MLP, while keeping the final GNN architecture unchanged for training on the resulting distilled dataset. Specifically, we consider GCN~\cite{kipf2016semi}, GraphSAGE~\cite{hamilton2017inductive}, and GIN~\cite{xu2018powerful} as the final GNN architectures.

As shown in Table~\ref{tab:innerloop}, replacing the inner-loop GNN with an MLP generally reduces performance. The drop is modest for GCN, whose computation becomes closer to an MLP when $\mathbf{A}'=\mathbf{I}$, but is more noticeable for GraphSAGE and GIN, whose parameterizations remain more different from a standard MLP. These results suggest that the inner-loop GNN serves an alignment role with the final LM--GNN training stage.

\vspace{4mm}
\section{Analysis of Structure-free Dataset Distillation}\label{sec:app:sfdd}

In this section, we analyze why the synthetic dataset can still reflect the structural information of the full dataset even when the synthetic adjacency matrix is fixed as $\mathbf{A}' = \mathbf{I}$. 

\subsection{Gradient Flow of Structural Information}
Based on the outer-loop KRR objective in Eq.~\eqref{eq:krr_obj}--\eqref{eq:krr_loss_labeled}, the gradient with respect to the synthetic representations $\mathbf{H}_S$ can be decomposed by the chain rule as
\[
\frac{\partial \mathcal{L}_{\mathrm{KRR}}}{\partial \mathbf{H}_\mathcal{S}}
=
\frac{\partial \mathcal{L}_{\mathrm{KRR}}}{\partial \mathbf{K}_\mathcal{TS}}
\frac{\partial \mathbf{K}_{\mathcal{TS}}}{\partial \mathbf{H}_\mathcal{S}}
+
\frac{\partial \mathcal{L}_{\mathrm{KRR}}}{\partial \mathbf{K}_\mathcal{SS}}
\frac{\partial \mathbf{K}_\mathcal{SS}}{\partial \mathbf{H}_\mathcal{S}}.
\]
Here, $\mathbf{H}_\mathcal{S}=[\mathbf{h}'_1,\mathbf{h}'_2,\ldots,\mathbf{h}'_M]^\top$, where $\mathbf{h}'_j$ denotes the representation of the $j$-th synthetic node. Among the two terms, the first term is key to explaining how structural information from the full dataset influences the synthetic dataset, since $\mathbf{K}_{\mathcal{TS}}=\mathbf{K}(\mathbf{H}_\mathcal{T},\mathbf{H}_\mathcal{S})$ is the cross-kernel matrix between the full dataset $\mathcal{T}$ and the synthetic dataset $\mathcal{S}$. Because $\mathbf{K}(\cdot,\cdot)$ is a graph-aware kernel, each entry of $\mathbf{K}_{\mathcal{TS}}$ is computed under the structural context of the original graph defined by $A$, rather than from feature similarity alone. More specifically, for each synthetic node representation $\mathbf{h}'_j$,
\[
\begin{aligned}
\frac{\partial \mathcal{L}_{\mathrm{KRR}}}{\partial \mathbf{h}'_j}
&=
\sum_i
\frac{\partial \mathcal{L}_{\mathrm{KRR}}}{\partial [\mathbf{K}_{\mathcal{TS}}]_{ij}}
\frac{\partial [\mathbf{K}_{\mathcal{TS}}]_{ij}}{\partial \mathbf{h}'_j} \\
&\quad+
\sum_{m,n}
\frac{\partial \mathcal{L}_{\mathrm{KRR}}}{\partial [\mathbf{K}_{\mathcal{SS}}]_{mn}}
\frac{\partial [\mathbf{K}_{\mathcal{SS}}]_{mn}}{\partial \mathbf{h}'_j}.
\end{aligned}
\]
This shows that each $\mathbf{h}'_j$ is updated according to its contribution to the labeled prediction loss through the kernel matrices. Therefore, even with $\mathbf{A}' = \mathbf{I}$, the synthetic representations are optimized through $\mathbf{K}_{\mathcal{TS}}$ to encode structural information from the full dataset that is useful for prediction.

\begin{table}[t]
\centering
\small
\setlength{\tabcolsep}{3.0pt}
\begin{tabular}{ccccc}
\toprule
Dataset & Ratio (\%) & Full ($=\mathbf{A}$) & 50\% & 100\% ($=\mathbf{I}$) \\
\midrule
Cora & 2.5 & \textbf{84.4} & 83.5 & 82.2 \\
Photo & 1.0 & \textbf{83.8} & 82.5 & 79.6 \\
Computers & 1.0 & \textbf{86.4} & 85.8 & 85.0 \\
\bottomrule
\end{tabular}
\caption{Effect of the original graph structure used in the outer-loop graph-aware kernel during distillation. The synthetic graph structure is fixed to $\mathbf{A}'=\mathbf{I}$ in all cases, while the original graph used during distillation is varied by edge dropping.}
\label{tab:edge_drop_kernel}
\end{table}

\subsection{Experiments}

To further examine whether structural information is incorporated even when the synthetic graph is fixed to $\mathbf{A}'=\mathbf{I}$, we conduct an additional experiment. During distillation, we keep the synthetic graph fixed as $\mathbf{A}'=\mathbf{I}$ and change only the original graph used in the outer-loop graph-aware kernel $\mathbf{K}_{\mathcal{TS}}$ during distillation. Specifically, we compare the full graph $\mathbf{A}$ with edge-dropped versions of $\mathbf{A}$ using 50\% and 100\% drop rates, while keeping the final training and inference protocol unchanged on the original full graph.

As shown in Table~\ref{tab:edge_drop_kernel}, using the full graph consistently performs best across all datasets, and performance decreases as more edges are removed. This suggests that, although \method does not explicitly synthesize edges, the original graph structure still affects the learned synthetic datasets through the graph-aware kernel and KRR objective.

\end{document}